\documentclass[runningheads]{llncs}
\usepackage{times}
\usepackage{amsmath}
\usepackage{amsfonts}
\usepackage{amssymb}
\usepackage{graphicx}
\usepackage{makeidx}
\usepackage{wrapfig}
\usepackage{hyperref}
\usepackage{caption}
\usepackage{natbib}
\usepackage[width=122mm,left=12mm,paperwidth=146mm,height=193mm,top=12mm,paperheight=217mm]{geometry}
\graphicspath{{./images}}

\begin{document}
\pagestyle{headings}
\mainmatter
\title{Large-scale Classification of Fine-Art Paintings: \\ Learning The Right Metric on The Right Feature} % Replace with your title

\titlerunning{Large-scale Classification of Fine-Art Paintings}

\authorrunning{Babak Saleh \and Ahmed Elgammal}

\author{Babak Saleh, Ahmed Elgammal}
\institute{Department of Computer Science \\ Rutgers University, NJ, USA \\ \href{mailto:babaks@cs.rutgers.edu}{\tt\ \{babaks,elgammal\}@cs.rutgers.edu}}

\maketitle
\begin{abstract}
In the past few years, the number of fine-art collections that are digitized and publicly available has been growing rapidly. With the availability of such large collections of digitized artworks comes the need to develop multimedia systems to archive and retrieve this pool of data. Measuring the visual similarity between artistic items is an essential step for such multimedia systems, which can benefit more high-level multimedia tasks. In order to model this similarity between paintings, we should extract the appropriate visual features for paintings and find out the best approach to learn the similarity metric based on these features.
We investigate a comprehensive list of visual features and metric learning approaches to learn an optimized similarity measure between paintings. We develop a machine that is able to make aesthetic-related semantic-level judgments, such as predicting a painting's style, genre, and artist, as well as providing similarity measures optimized based on the knowledge available in the domain of art historical interpretation.
Our experiments show the value of using this similarity measure for the aforementioned prediction tasks. 
%
%Automatic analysis of a collection of fine-art paintings is an important task that can benefit from artificial intelligence. The process of analyzing a painting involves extracting some artistic knowledge, known as elements of art. Majority of this knowledge is extracted visually and can be automated via computer vision algorithms. However majority of visual features and learning algorithms in the field of computer vision are designed for natural photographic images rather than artistic paintings. 
%In this work we delve into the analysis of paintings by classification of their style, genre and painter. Toward this goal we investigate a variety of metric learning approaches and visual features to properly model the notion of similarity between paintings. We have performed an extensive comparative study to find most accurate models for classification of \textit{Style}, \textit{Genre} and \textit{Artist} of fine-art paintings.
\end{abstract}

%A category including the fourth, optional field follows...
%\category{D.2.8}{Software Engineering}{Metrics}[complexity measures, performance measures]


%\keywords{ACM proceedings, \LaTeX, text tagging}
%
%\section{Introduction}
%\label{Sec:intro}
%\input{MM_Intro}
\section{Introduction}
\label{Sec:intro}
In the past few years, the number of fine-art collections that are digitized and publicly available has been growing rapidly. Such collections span classical~\footnote{ Examples:  \href{http://www.wikiart.org/en/}{ Wikiart};~\href{http://arkyves.org/}{Arkyves};~\href{http://www.bbc.co.uk/arts/yourpaintings/}{BBC Yourpainting}} and modern and contemporary artworks~\footnote{Examples: \href{https://www.artsy.net/}{Artsy};~\href{https://www.behance.net/}{Behance};~\href{https://www.artnet.com/}{Artnet}}.  With the availability of such large collections of digitized artworks comes the need to develop multimedia systems to archive and retrieve this pool of data. Typically these collections, in particular early modern ones, come with metadata in the form of annotations by art historians and curators, including information about each painting's artist, style, date, genre, etc. For online galleries displaying contemporary artwork, there is a need to develop automated recommendation systems that can retrieve ``similar" paintings that the user might like to buy. This highlights the need to investigate metrics of visual similarity among digitized paintings that are optimized for the domain of painting.

\begin{figure}[t]
\begin{center}
\includegraphics[scale=.45]{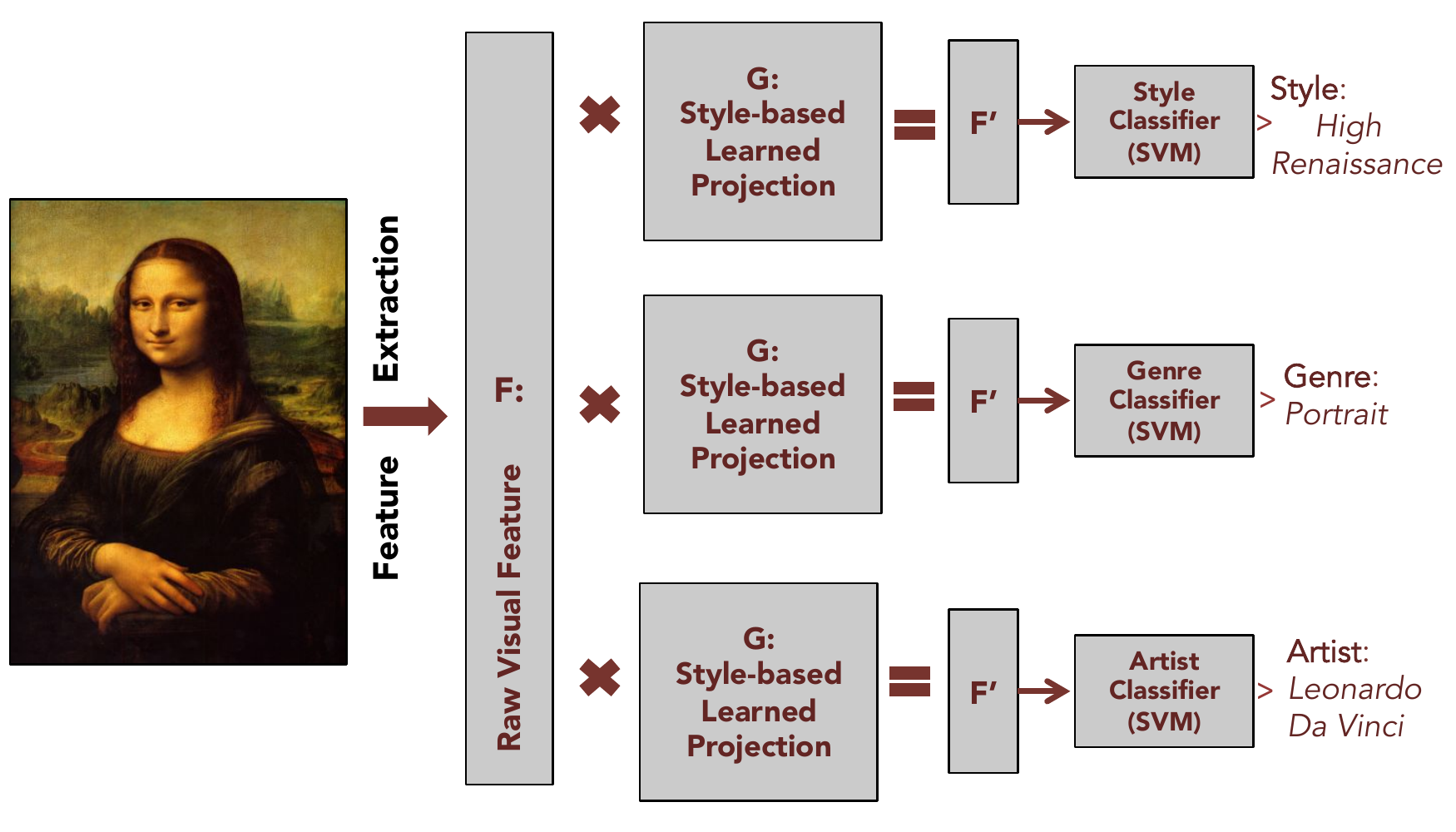}
\caption{Illustration of our system for classification of fine-art paintings. We investigated variety of visual features and metric learning approaches to recognize \textit{Style, Genre} and \textit{Artist} of a painting.}
\label{fig:method1}
\end{center}
\vspace{-20pt}
\end{figure}

The field of computer vision has made significant leaps in getting digital systems to recognize and categorize objects and scenes in images and videos. These advances have been driven by a wide spread need for the technology, since cameras are everywhere now. However a person looking at a painting can make sophisticated inferences beyond just recognizing a tree, a chair, or the figure of Christ. Even individuals without specific art historical training can make assumptions about a painting's genre (portrait or landscape), its style (impressionist or abstract), what century it was created, the artists who likely created the work and so on. Obviously, the accuracy of such assumptions depends on the viewer's level of knowledge and exposure to art history. Learning and judging such complex visual concepts is an impressive ability of human perception~\cite{arnheim1969visual}.

The ultimate goal of our research is to develop a machine that is able to make aesthetic-related semantic-level judgments, such as predicting a painting's style, genre, and artist, as well as providing similarity measures optimized based on the knowledge available in the domain of art historical interpretation. Immediate questions that arise include, but are not limited to: What visual features should be used to encode information in images of paintings? How does one weigh different visual features to achieve a useful similarity measure? What type of art historical knowledge should be used to optimize such similarity measures? In this paper we address these questions and aim to provide answers that can benefit researchers in the area of computer-based analysis of art. Our work is based on a systematic methodology and a comprehensive evaluation on one of the largest available digitized art datasets.

Artists use different concepts to describe paintings. In particular, stylistic elements, such as space, texture, form, shape, color, tone and line are used. Other principles include movement, unity, harmony, variety, balance, contrast, proportion, and pattern. To this might be added physical attributes, like brush strokes as well as subject matter and other descriptive concepts~\cite{lois}. 
%We call these concepts collectively artistic concepts.

For the task of computer analyses of art, researchers have engineered and investigated various visual features\footnote{In contrast to art disciplines, in the fields of computer vision and machine learning, researchers use the term``visual features" to denote statistical measurements that are extracted from images for the task of classification. In this paper we stick to this typical terminology.} that encode some of these artistic concepts, in particular brush strokes and color, which are encoded as low-level features such as texture statistics and color histograms (e.g.~\cite{ Jia12, Lombardi}). Color and texture are highly prone to variations during the digitization of paintings; color is also affected by a painting's age. The effect of digitization on the computational analysis of paintings is investigated in great depth by Polatkan et al.~\cite{brdahujapo09}. This highlights the need to carefully design visual features that are suitable for the analysis of paintings.

\begin{table*}[t]
\begin{center}
\includegraphics[width=.95\textwidth, height=.3\textheight]{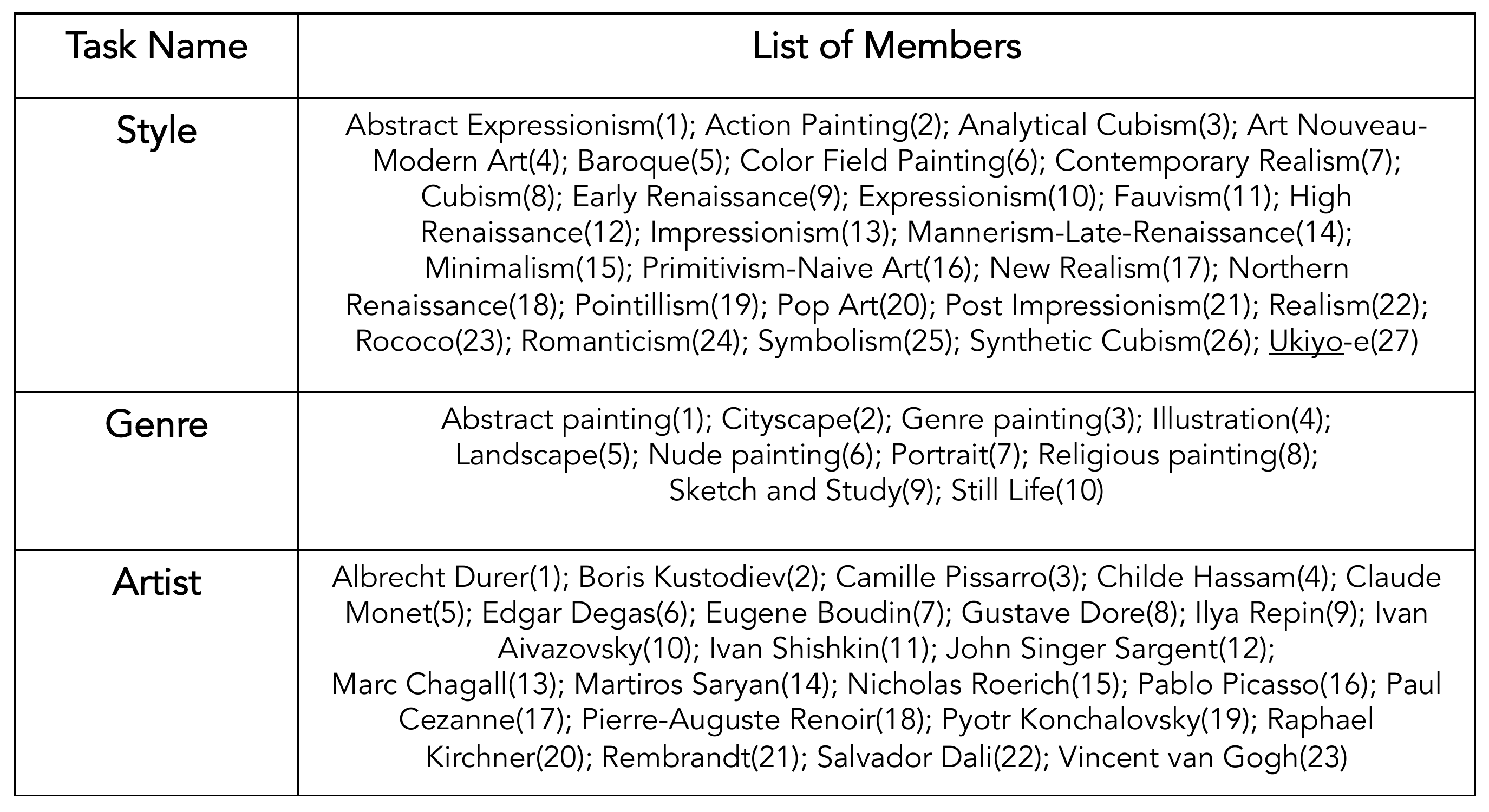}
\caption{List of Styles, Genres and Artists in our collection of fine-art paintings. Numbers in the parenthesis are index of the row/column in confusion matrices \ref{fig:conf_style_new}, \ref{fig:conf_genre_new}$\&$ \ref{fig:conf_artist_new} accordingly.}
\label{fig:table_names}
\vspace{-30pt}
\end{center}
\end{table*}

Clearly, it would be a cumbersome process to engineer visual features that encode all the aforementioned artistic concepts. Recent advances in computer vision, using deep neural networks, showed the advantage of ``learning" the features from data instead of engineering such features.  However, It would also be impractical to learn visual features that encode such artistic concepts, since that would require extensive annotation of these concepts in each image within a large training and testing dataset. Obtaining such annotations require expertise in the field of art history that can not be achieved with typical crowed-sourcing annotators.  

Given the aforementioned challenges to engineering or learning suitable visual features for painting, in this paper we follow an alternative strategy. We mainly investigate different state-of-the-art visual elements, ranging from low-level elements to semantic-level elements. We then use metric learning to achieve optimal similarity metrics between paintings that are optimized for specific prediction tasks, namely style, genre, and artist classification. We chose these tasks to optimize and evaluate the metrics since, ultimately, the goal of any art recommendation system would be to retrieve artworks that are similar along the directions of these high-level semantic concepts. Moreover, annotations for these tasks are widely available and more often agreed-upon by art historians and critics, which facilitates training and testing the metrics.

In this paper we investigate a large space of visual features and learning methodologies for the aforementioned prediction tasks. We propose and compare three learning methodologies to optimize such tasks. We present results of a comprehensive comparative study that spans four state-of-the-art visual features, five metric learning approaches and the proposed three learning methodologies, evaluated on the aforementioned three artistic prediction tasks.

\section{Related Work}
\label{Sec:rel}
\label{Sec:rel}
On the subject of painting, computers have been used for a diverse set of tasks. Traditionally, image processing techniques have been used to provide art historians with quantification tools, such as pigmentation analysis, statistical quantification of brush strokes, etc. We refer the reader to \cite{stork2009computer,stork_book} for comprehensive surveys on this subject. 

Several studies have addressed the question of which features should be used to encode information in paintings. Most of the research concerning the classification of paintings utilizes low-level features encoding color, shadow, texture, and edges. For example Lombardi~\cite{Lombardi} has presented a study of the performance of these types of features for the task of artist classification among a small set of artists using several supervised and unsupervised learning methodologies. In that paper the style of the painting was identified as a result of recognizing the artist.
 
Since brushstrokes provide a signature that can help identify the artist, designing visual features that encode brushstrokes has been widely adapted.(e.g.~\cite{sablatnig,li2004studying,lyu2004digital,johnson2008image,berezhnoy2009automatic,Jia12}). Typically, texture statistics are used for that purpose. However, as mentioned earlier, texture features are highly affected by the digitization resolution. Researchers also investigated the use of features based on local edge orientation histograms, such as SIFT~\cite{SIFT} and HOG~\cite{HOG}. For example,~\cite{khan} used SIFT features within a Bag-of-words pipeline to discriminate among a set of eight artists.  

Arora et al.~\cite{Arora12} presented a comparative study for the task of style classification, which evaluated low-level features, such as SIFT and Color SIFT~\cite{csift}, versus semantic-level features, namely Classemes~\cite{aleb}, which encodes object presence in the image. It was found that semantic-level features significantly outperform low-level features for this task. However the evaluation was conducted on a small dataset of 7 styles, with 70 paintings in each style. Carneiro et al~\cite{Carneiro12} also concluded that low-level texture and color features are not effective because of inconsistent color and texture patterns that describe the visual classes in paintings.

More recently, Saleh et al~\cite{saleh2014knowledge} used metric learning approaches for finding influence paths between painters based on their paintings. They evaluated three metric learning approaches to optimize a metric over low-level HOG features. In contrast to that work, the evaluation presented in this paper is much wider in scope since we address three tasks (style, genre and artist prediction), we cover features spanning from low-level to semantic-level and we evaluate five metric learning approaches. Moreover, The dataset of~\cite{saleh2014knowledge} has only 1710 images from 66 artists, while we conducted our experiments on 81,449 images painted by 1119 artists. Bar et al~\cite{israel} proposed an approach for style classification based on features obtained from a convolution neural network pre-trained on an image categorization task. In contrast we show that we can achieve better results with much lower dimensional features that are directly optimized for style and genre classification. Lower dimensionality of the features is preferred for indexing large image collections.

\section{Methodology}
\label{Sec:model}
\label{Sec:model}
In this section we explain the methodology that we follow to find the most appropriate combination of visual features and metrics that produce accurate similarity measurements. We acquire these measurements to mimic the art historian's ability to categorize paintings based on their style, genre and the artist who made it. In the first step, we extract visual features from the image. These visual features range from low-level (e.g. edges) to high-level (e.g. objects in the painting). More importantly, in the next step we learn how to adjust these features for different classification tasks by learning the appropriate metrics. Given the learned metric we are able to project paintings from a high dimensional space of raw visual information to a meaningful space with much lower dimensionality. Additionally, learning a classifier in this low-dimensional space can be easily scaled up for large collections. 

In the rest of this section: First, we introduce our collection of fine-art paintings and explain what are the tasks that we target in this work. Later, we explore methodologies that we consider in this work to find the most accurate system for aforementioned tasks. Finally, we explain different types of visual features that we use to represent images of paintings and discuss metric learning approaches that we applied to find the proper notion of similarity between paintings.

\begin{figure}[t]
\begin{center}
\includegraphics[scale=.45]{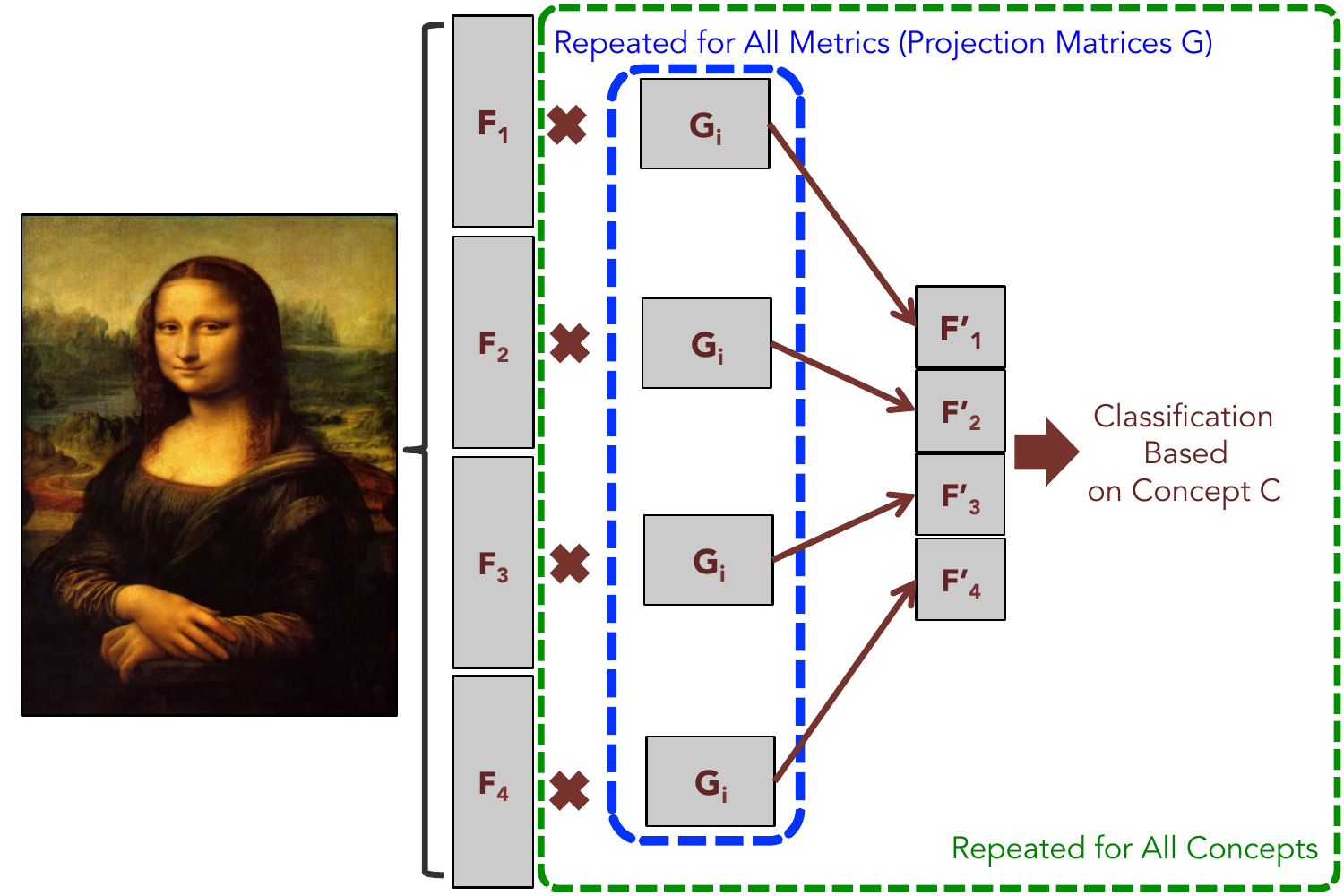}
\caption{Illustration of our second methodology - Feature Fusion.}
\label{fig:method2}
\vspace{-15pt}
\end{center}
\end{figure}

\subsection{Dataset and Proposed Tasks}
In order to gather our collection of fine-art paintings, we used the publicly available dataset of \textit{"Wikiart paintings"}\footnote{http://www.wikiart.org/}; which, to the best of our knowledge, is the largest online public collection of digitized artworks. This collection has images of 81,449 fine-art paintings from 1,119 artists ranging from fifteen centuries to contemporary artists. These paintings are from 27 different styles (Abstract, Byzantine, Baroque, etc.) and 45 different genres (Interior, Landscape, etc.) Previous work~\citep{saleh2014knowledge,Carneiro12} used different resources and made smaller collections with limited variability in terms of style, genre and artists. The work of ~\cite{israel} is the closest to our work in terms of data collection procedure, but the number of images in their collection is half of ours.

We target automatic classification of paintings based on their style, genre and artist using visual features that are automatically extracted using computer vision algorithms. Each of these tasks has its own challenges and limitations. For example, there are large variations in terms of visual appearances in paintings from one specific style. However, this variation is much more limited for paintings by one artist. These larger intra-class variations suggests that style classification based on visual features is more challenging than artist classification. For each of the tasks we selected a subset of the data that ensure enough samples for training and testing. In particular for style classification  we use a subset of the date with 27 styles where each style has at least 1500 paintings with no restriction on genre or artists, with a total of 78,449  images. For genre classification we use a subset with 10 genre classes, where each genre has at least 1500 paintings with no restriction of style or genre, with a total of 63,691 images. Similarly for artist classification we use a subset of 23 artists, where each of them has at least 500 paintings, with a total of 18,599 images.
 Table~\ref{fig:table_names} lists the set of style, genre, and artist labels.

%Previous work~\citep{saleh2014knowledge,Carneiro12} used different resources and made smaller collections with limited variability in terms of style, genre and artists. The work of ~\cite{israel} is the closest to our work in terms of data collection procedure, but the number of images in their collection is half of ours. 

\subsection{Classification Methodology}
In order to classify paintings based on their style, genre or artist we followed three methodologies. 

{\em Metric Learning:}
First, as depicted in figure~\ref{fig:method1}, we extract visual features from images of paintings. 
For each of these prediction tasks, we learn a similarity metric optimized for it, i.e. style-optimized metric, genre-optimized metric and artist-optimized metric. Each metric induces a projector to a corresponding feature space optimized for the corresponding task.  
Having the metric learned, we project the raw visual features into the new optimized feature space and learn classifiers for the corresponding prediction task. For that purpose we learn a set of one-vs-all SVM classifiers for each of the labels in table~\ref{fig:table_names} for each of the tasks.

While our first strategy focuses on classification based on combinations of a metric and a visual feature, the next two methodologies that we followed fuse different features or different metrics. 

{\em Feature fusion:}
The second methodology that we used for classification is depicted in figure~\ref{fig:method2}. In this case, we extract different types of visual features (four types of features as will explained next). Based on the prediction task (e.g. style) we learn the metric for each type of feature as before. After projecting these features separately, we concatenate them to make the final feature vector. The classification will be based on training classifiers using these final features. This feature fusion is important as we want to capture different types of visual information by using different types of features. Also concatenating all features together and learn a metric on top of this huge feature vector is computationally intractable. Because of this issue, we learn metrics on feature separately and after projecting features by these metrics, we can concatenate them for classification purposes. 

\begin{figure}[t]
\begin{center}
\includegraphics[scale=.45]{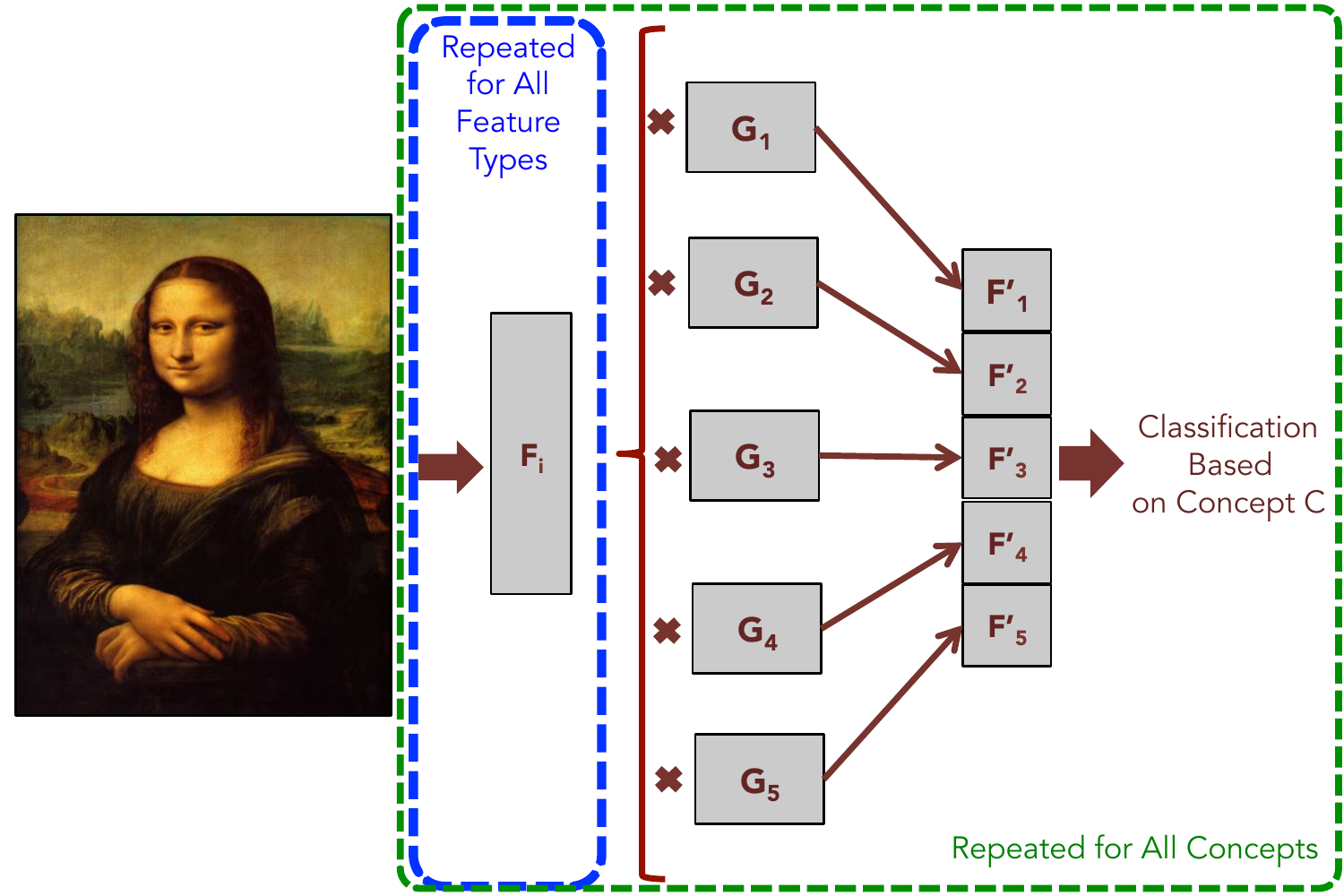}
\caption{Illustration of our third methodology-- Metric Fusion.}
\label{fig:method3}
\end{center}
\vspace{-15pt}
\end{figure}

{\em Metric-fusion:}
The third methodology (figure~\ref{fig:method3}) projects each visual features using multiple metrics (in our experiment we used five metrics as will be explained next) and then fuses the resulting optimized feature spaces to obtain a final feature vector for classification. This is an important strategy, because each one of the metric learning approaches use a different criteria to learn the similarity measurement. By learning all metrics individually (on the same type of feature), we make sure that we took into account all criteria (e.g. information theory along with neighbor hood analysis).

%The third methodology (figure~\ref{fig:method3}) fuses projected feature vectors before training classifiers. However, in this scenario we fix the type of feature first. Learn all different similarity metrics (in our experiment we tried five metrics) and project the feature based on these individual metrics and concatenate the outputs.

\subsection{Visual Features}
Visual features in computer vision literature are either engineered and extracted in an unsupervised way (e.g. HOG, GIST) or learned based on optimizing a specific task, typically categorization of objects or scenes (e.g. CNN-based features).  This results in high-dimensional feature vectors that might not necessary correspond to nameable (semantic-level) characteristics of an image. Based on the ability to find a meaning, visual features can be categorized into low-level and high-level. Low-level features are visual descriptors that there is no explicit meaning for each dimension of them, while high-level visual features are designed to capture some notions (usually objects). For this work, we investigated some state-of-the-art representatives of these two categories:
 
{\em Low-level Features:} On one hand, in order to capture low-level visual information we extracted GIST features~\citep{GIST}, which are holistic features that are designed for scene categorization.  GIST features provide a 512 real-valued representation that implicitly captures the dominant spatial structure of the image.

%\cite{saleh2014toward} showed that GIST outperforms other low-level visual features (e.g. SIFT) for different classification tasks. 

{\em Learned Semantic-level  Features:} On the other hand, for the purpose of semantic representation of the images, we extracted three object-based representation of the images: Classeme~\citep{aleb}, Picodes~\citep{picodes}, and CNN-based features~\citep{alex}. In all these three features, each element of the feature vector represents the confidence of the presence of an object-category in the image, therefore they provide a semantic encoding of the images. However, for learning these features, the object-categories are generic and are not art-specific.  First two features are designed to capture the presence of a set of basic-level object categories as following: a list of entry-level categories (e.g. horse and cross) is used for downloading a large collection of images from the web. For each image a comprehensive set of low-level visual features are extracted and one classifier is learned for each category. For a given test image, these classifiers are applied on the image and the responses (confidences) make the final feature vector. We followed the implementation of~\citep{classemes} and for each image extracted a 2659 dimensional real-valued Classeme feature vector and a 2048 dimensional binary-value Picodes feature. 

Convolutional Neural Networks(CNN)~\citep{lecun1998gradient} showed a remarkable performance for the task of large-scale image categorization~\citep{alex}. CNNs have four convolutional layers followed by three fully connected layers. Bar et al~\cite{israel} showed that a combination of the output of these fully connected layers achieve a superior performance for the task of style classification of paintings. Following this observation we used the last layer of a pre-trained CNN~\citep{alex} (1000 dimensional real-valued vectors) as another feature vector.

\begin{figure}[t]
%\begin{center}
\centering
\includegraphics[width=.7\textwidth]{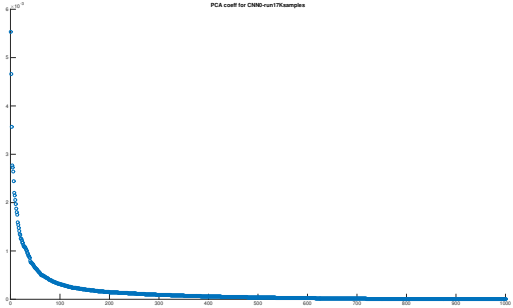}
\caption{PCA coefficients for CNN features}
%\end{center}
\label{fig:pca}
\vspace{-15pt}
\end{figure}

\subsection{Metric Learning}
The purpose of Metric Learning is to find some pair-wise real-valued function $d_M(x,x')$ which is non-negative, symmetric, obeys the triangle inequality and returns zero if and only if $x$ and $x'$ are the same point. Training such a function in a general form can be seen as the following optimization problem:
\begin{equation}
\label{Eq:metric_learning}
  \min_M l(M,D) + \lambda R(M)
\end{equation}
This optimization has two sides, first it tries to minimize the amount of loss $l(M,D)$ by using metric $M$ over data samples $D$ while trying to adjust the model by the regularization term $R(M)$. The first term shows the accuracy of the trained metric and second one estimates its capability over new data and avoids overfitting.
Based on the enforced constraints, the resulted metric can be linear or non-linear and depending on the amount of labels used for training, it can be supervised or unsupervised. 

For consistency over the metric learning algorithms, we need to fix the notation first. We learn the matrix $M$ that will be used in Generalized Mahalanobis Distance: $d_M(x,x') = \sqrt{(x-x')' M (x-x')}$, where $M$ by definition is a positive semi-definite matrix and can be decomposed as $M = G^{T}G$. We use this matrix $G$ to project raw visual features. Measuring similarity in this projection space is simply computing the euclidean distance between two item.

It is interesting that we can reduce the dimension of features during learning the metric when $M$ is a low rank matrix. More importantly, there are significantly important information in the ground truth annotation associated with paintings that we use to learn a more reliable metric in a supervised fashion for both the linear and non-linear cases. We consider following approaches that differ based on the form of $M$ or the amount of regularization.

\subsubsection{Neighborhood Component Analysis (NCA)}
The objective function of NCA~\citep{NCA} is related to analyzing the nearest neighbors. The idea starts with projecting the data by matrix $M$ and training a leave-one-out classifier. Then the probability of correctly classifying $x_i$ is 
$P_i = \sum_{j:y_i = y_j}P_{ij}$, where $P_{ij}$ is the mean expected loss of classifying $x_i$ as a member of class $j$.

Then this metric is learned by optimizing the following term: $\max_{M} \sum_i P_i$.
We can decompose $M$ as $L'*L$ and choosing a rectangular $L$ will result in a low-rank matrix $M$. 
Although this method is easy to understand and implement, it is subject to local minimums. This happens due to the non-convexity of the proposed optimization problem. The next approach has the advantage of solving a convex optimization.

\begin{figure}[t]
%\begin{center}
\centering
\includegraphics[width=.6\textwidth]{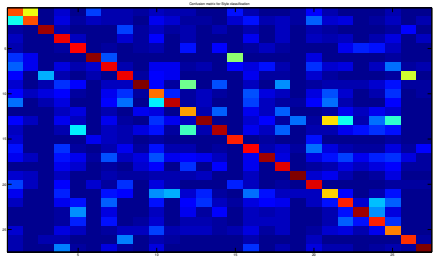}
\caption{Confusion matrix for Style classification. Confusions are meaningful only when seen in color.}
\label{fig:conf_style_new}
\vspace{-10pt}
%\end{center}
\end{figure}

\subsubsection{Large Margin Nearest Neighbors (LMNN)}
LMNN~\citep{LMNN} is an approach for learning a Mahalanobis distance, which is widely used because of its global optimum solution and superior performance in practice. The learning of this metric involves a set of constrains, all of which are defined locally. This means that LMNN enforces the $k$ nearest neighbor of any training instance belonging to the same class (these instances are called ``target neighbors"). This should be done while all the instances of other classes, referred as ``impostors", should be far from this point. For finding the target neighbors, Euclidean distance has been applied to each pair of samples, resulting in the following formulation:

\begin{eqnarray*}
\min_{M}(1-\mu)\sum_{(x_i,x_j)\in T} d_M^2 (x_i,x_j) + \mu\sum_{i,j,k}\eta_{i,j,k} \nonumber \\
s.t. :  d_{M}^{2}(x_i,x_k) - d_{M}^{2}(x_i,x_j) \geq 1-\eta_{i,j,k}  \forall(x_i,x_j,x_k) \in I. 
\end{eqnarray*} 

Where $T$ stands for the set of \textit{Target} neighbors and $I$ represents \textit{Impostors}. Since these constrains are locally defined, this optimization leads to a convex formulation and a global solution. 
This metric learning approach is related to Support Vector Machines (SVM) in principle, which theoretically engages its usage along with SVM for the task of classification. 

Due to the popularity of LMNN, different variations of it have been introduced, including a non-linear version called gb-LMNN~\citep{LMNN} which we used in our experiments as well. However its performance for classification tasks was worse that linear LMNN. We assume this poor performance is rooted in the nature of visual features that we extract for paintings.

\begin{figure}[t]
%\begin{center}
\centering
\includegraphics[width=.6\textwidth]{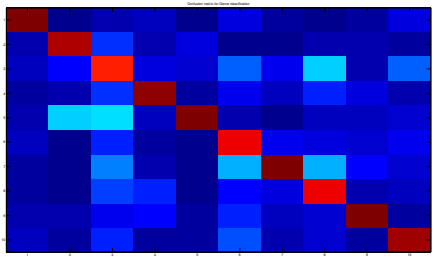}
\caption{Confusion matrix for Genre classification. Confusions are meaningful only when seen in color.}
\label{fig:conf_genre_new}
\vspace{-10pt}
%\end{center}
\end{figure}
  
\subsubsection{Boost Metric}
This approach is based on the fact that a positive semi-definite matrix can be decomposed into a linear combination of trace-one rank-one matrices. Shen et al~\cite{boostmetric} use this fact and instead of learning $M$, finds a set of weaker metrics that can be combined and  give the final metric. They treat each of these matrices as a \textit{Weak Learner}, which is used in the literature of Boosting methods. The resulting algorithm applies the idea of AdaBoost to Mahalanobis distance, which has been shown to be quiet efficient in practice.

This method is particularly of our interest, because we can learn an individual metric for each style of paintings and finally merge these metrics to get a unique final metric. Theoretically the final metric can perform well to find similarities inside each style/genre of paintings as well. 

\subsubsection{Information Theory Metric Learning (ITML)}
This metric learning algorithm is based on Information theory rather than Mahalanobis distances. In other words the optimization problem of learning a metric involves an information measure.
 
Davis et al~\cite{ITML} introduce the measure of \textit{LogDet divergence regularization} between two matrices $M, M'$(can be interpreted as metrics). 
By using this measure, learning the metric can be represented by: 

\begin{eqnarray*}
\min_{M' \in PSD} D_{ld}(M,M') + \gamma \sum_{i,j} \epsilon_{i,j} \nonumber \\
s.t. :  d_{M'}^{2}(x_i,x_j) \leq u+\epsilon_{i,j}    \forall(x_i,x_j) \in S. \nonumber \\
  d_{M'}^{2}(x_i,x_j) \geq v-\epsilon_{i,j}    \forall(x_i,x_j) \in D.
\end{eqnarray*} 

Learning ITML via this formulation aims to satisfy a set of \textit{Similarity(S)} and \textit{Dissimilarity(D)} constrains while keeping the new metric $M'$ close to the initial metric $M$. There are two key features of the \textit{LogDet divergence}: 1) It is finite if and only if matrices are positive semi-definite(PSD),  2) This function is rank-preserving. 

These properties indicate that if we start learning the metric $M'$ by setting the initial matrix $M$ as identity matrix($I$), ITML returns a metric that is from the same rank and is very similar to the Euclidean distance. 

Although this iterative process converges to a global minimum which performs well in practice, it is very sensitive to the choice of initialization of metric($M$).

\subsubsection{Metric Learning for Kernel Regression (MLKR)}
Similar to NCA objective function, which minimizes the classification error; Weinberger and Tesauro~\cite{MLKR} learn a metric by optimizing the leave-one-out error for the task of kernel regression. In kernel regression, there is an essential need for proper distances between points that will be used for weighting sample data. MLKR learn this distance by minimizing the leave-one-out error for regression on training data. Although this metric learning method is designed for kernel regression, the resulted distance function can be used in variety of tasks.

\section{Experiments}
\label{Sec:Exp}
\label{Sec:Exp}
%In this section we explain our experiments by first introducing the dataset and the visual features that we used in our comparative study. Next we  explain the way we learn the metrics mentioned in Sec.\ref{Sec:model} to capture more meaningful similarity between paintings. Finally we conduct a series of experiments to show our performance for three tasks of \textit{Style, Genre} and \textit{Artist classification} by applying these new measure of similarity that is computed by the learned metrics.

\subsection{Experimental Setting}
%\subsubsection{Dataset}
%We used the publicly available dataset of \textit{"Wikiart paintings"}\footnote{http://www.wikiart.org/}; which, to the best of our knowledge, is the largest online public collection of artworks. This collection has images of 81,449 fine-art paintings from 1,119 artist. These paintings are from 27 different styles (Abstract, Byzantine, Baroque, etc.) and 45 different genres (Interior, Landscape, Mosaic, etc.) Previous work~\citep{saleh2014knowledge,Carneiro12} used different resources and made smaller collections with limited variability in terms of style, genre and artists. The work of ~\cite{israel} is the closest to our work in terms of data collection procedure, but the number of images in their collection is half of ours. 

\subsubsection{Visual Features}
As we explained in section~\ref{Sec:model}, we extract GIST features as low-level visual features and Classeme, Picodes and CNN-based features as the high-level semantic features. We followed the original implementation of Oliva and Torralba~\cite{GIST} to get a 512 dimensional feature vector. For Classeme and Picodes we used the implementation of Bergamo et al~\cite{aleb}, resulting in 2659 dimensional Classeme features and 2048 dimensional Picodes features. We used the implementation of Vedaldi and Lenc~\cite{matconv} to extract 1000 dimensional feature vectors of the last layer of CNN.

\begin{figure}[t]
%\begin{center}
\centering
\includegraphics[width=.6\textwidth]{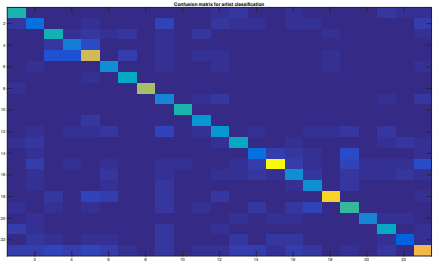}
\caption{Confusion matrix for Artist classification. Confusions are meaningful only when seen in color.}
\label{fig:conf_artist_new}
\vspace{-10pt}
%\end{center}
\end{figure}

Object-based representations of the images produce feature vectors that are much higher in dimensionality than GIST descriptors. In the sake of a fair comparison of all types of features for the task of metric learning, we transformed all feature vectors to have the same size as GIST (512 dimensional). We did this by applying Principle Component Analysis (PCA) for each type and projecting the original features onto the first 512 eigenvectors (with biggest eigenvalues). In order to verify the quality of projection, we looked at the corresponding coefficients of eigenvalues for PCA projections. Independent of feature type, the value of these coefficients drops significantly after the first 500 eigenvectors. For example, figure~\ref{fig:pca} plots these coefficients of PCA projection for CNN features. Summation of the first 500 coefficients is $95.88 \%$ of the total summation. This shows that our projections (with 512 eigenvectors) captures the true underlying space of the original features. Using these reduced features speeds up the metric learning process as well.

\begin{table}[t]
\begin{center}
\begin{tabular}{|l|l|l|l|l|l|}
\hline
 Metric~/~Features & GIST & Classemes & Picodes & CNN & Dim.\\
 \hline \hline
  Baseline & 10.83 & 22.62 & 20.76 & 12.32 & 512\\
  Boost & \textbf{16.07} & \textbf{31.77} & \textbf{28.58} & 15.18 & 512\\
  ITML & 13.02 & 30.67 & 28.42 & 15.34 & 512\\
  LMNN & 12.54 & 27 & 24.14 & \textbf{16.83} & 100\\
  MLKR & 12.65 & 24.12 & 14.86 & 12.63 & 512\\
  NCA & 13.29 & 28.19 & 24.84 & 16.37 & 27\\
\hline 
\end{tabular}
\caption{Accuracy for the task of style classification}
\label{tab:style}
\end{center}
\vspace*{-30pt}
\end{table}

\subsubsection{Metric Learning}
We used implementation of~\cite{LMNN} to learn LMNN metric(both version of linear and non-linear) and MLKR~ \footnote{\href{http://www.cse.wustl.edu/~kilian/code/code.html}{http://www.cse.wustl.edu/~kilian/index.html}}. For the BoostMetric we slightly adjusted the implementation of~\cite{boostmetric}. For NCA we adopted its implementation by Fowlkes\footnote{\href{http://www.ics.uci.edu/~fowlkes/software/nca/}{http://www.ics.uci.edu/~fowlkes/}} to work on large scale feature vectors smoothly. For the case of ITML metric learning, we followed the original implementation of authors with the default setting. For the rest of methods, parameters are chosen through a grid search that finds the minimum nearest neighbor classification. Regarding the training time, learning the ITML metric was the fastest and learning NCA and LMNN were the slowest ones. Due to computational constrains we set the parameters of LMNN metric to reduce the size of features to 100. NCA metric reduces the dimension of features to the number of categories for each tasks: 27 for style classification, 23 for artist classification and 10 for genre classification. We randomly picked 3000 samples, which we used for metric learning. These samples follow the same distribution as original data and are not used for classification experiments.

\subsection{Classification Experiments}
For the purpose of metric learning, we conducted experiments with labels for three different tasks of style, genre and artist prediction. In following sections we investigate the performance of these metrics on different features for classification of aforementioned concepts.

We learned all the metrics in section\ref{Sec:model} for all 27 styles of paintings in our dataset (e.g. Expressionism, Realism, etc.). However, we did not use all the genres for learning metrics. In fact in our dataset we have 45 genres, some of which have less than 20 images. This makes the metric learning impractical and highly biased toward genres with larger number of paintings. Because of this issue, we focus on 10 genres with more than 1500 paintings. These genres are listed in table~\ref{fig:table_names}.
In all experiments we conducted 3 fold cross validation and reported the average accuracy over all partitions. We found the best value for penalty term in SVM (which is equal to 10) by three fold cross validation. In the next three sections, we explain settings and findings for each task independently. 

%
%\begin{table}[t]
%\begin{center}
%\begin{tabular}{|l|l|l|l|l|l|}
%\hline
% Metric~/~Features & GIST & Classemes & Picodes & CNN & Dim.\\
% \hline \hline
%  Baseline & 10.83 & 22.62 & 20.76 & 12.32 & 512\\
%  Boost & \textbf{16.07} & \textbf{31.77} & \textbf{28.58} & 15.18 & 512\\
%  ITML & 13.02 & 30.67 & 28.42 & 15.34 & 512\\
%  LMNN & 12.54 & 27 & 24.14 & \textbf{16.83} & 100\\
%  MLKR & 12.65 & 24.12 & 14.86 & 12.63 & 512\\
%  NCA & 13.29 & 28.19 & 24.84 & 16.37 & 27\\
%\hline 
%\end{tabular}
%\caption{Accuracy for the task of style classification}
%\label{tab:style}
%\end{center}
%\vspace{-20pt}
%\end{table}

\begin{table}[t]
\begin{center}
\begin{tabular}{|l|l|l|l|l|l|}
\hline
 Metric~/~Features & GIST& Classemes & Picodes & CNN & Dim.\\
 \hline \hline
  Baseline & 28.10 & 49.98 & 49.63 & 35.14 & 512\\
  Boost & 31.01  & \textbf{57.87} & \textbf{57.35} & 46.14 & 512\\
  ITML & 33.10 & 57.86 & 57.28 & 46.80 & 512\\
  LMNN & \textbf{39.06} & 54.96 & 54.42 & \textbf{49.98} & 100\\
  MLKR & 32.81 & 54.29 & 42.79 & 45.02 & 512\\
  NCA & 30.39 & 51.38 & 52.74 & 49.26 & 10\\
\hline 
\end{tabular}
\caption{Accuracy for the task of genre classification}
\label{tab:genre}
\end{center}
\vspace*{-30pt}
\end{table}

\subsubsection{Style Classification}
Table~\ref{tab:style} contains the result (accuracy percentage) of style classification (SVM) after applying different metrics on a set of features. Columns correspond to different features and rows are different metrics that are used for projecting features before learning style classifiers. In order to quantify the improvement by learning similarity metrics, we conducted a baseline experiment (first row in the table) as the following: For each type of features, we learn a set of one-vs-all classifiers on raw feature vectors. Generally Boost metric learning and ITML approaches give the highest in accuracy for the task of style classification over different visual features. However the greatest improvement over the baseline is gained by application of Boost metric on Classeme features. We visualized the confusion matrix for the task of style classification, when we learn Boost metric on Classeme features. 

Figure~\ref{fig:conf_style_new} shows this matrix, where red  represents higher values. Further analysis of some confusions that are captured in this matrix result in interesting findings. In the rest of this paragraph we explain some of these cases. First, we found that there is a big confusion between ``Abstract expressionism" (first row) and ``Action paintings" (second column). Art historians verify the fact that this confusion is meaningful and somehow expected. ``Action painting" is a type or subgenre of ``abstract expressionism" and are characterized by paintings created through a much more active process-- drips, flung paint, stepping on the canvas.

%``Abstract expressionism" is characterized by being messy paintings with applications of energetic colors, absence of recognizable subjects and small brushstroke.  Similarly ``Action paintings" is characterized by large paintings that are shapeless and flat, usually drawn by dripping brilliant color paints on canvas. 

%
%``Fauvism" is an early form of ``Expressionism", where both share the following characteristics: distortion, strong outlines with simple compositions and are free of brushstrokes. These styles use bold and pure colors and ``Fauvism" paintings are not realistic. 
%
%``Romanticism" (row 24) and ``Symbolism"(column 25) are confused as both styles are not real or natural and in fact Symbolism is considered as Pre-Romanticism. ``Realism" (column 22) and ``Rococo" (row 23) are confusing as well, since both are not idealized and there is no straightforward figures.
Another confusion happens between ``Expressionism" (column 10) and ``Fauvism" (row 11), which is actually expected based on art history literature. 
 ``Mannerism" (row 14) is a style of art during the (late)``Renaissance" (column 12), where they show unusual effect in scale and are less naturalistic than ``Early Renaissance". This similarity  between ``Mannerism" (row 14) and ``Renaissance" (column 12) is captured by our system as well where results in confusion during style classification. ``Minimalism" (column 15) and ``Color field paintings"(6th row) are mostly confused with each other. We can agree on this finding as we look at members of these styles and figure out the similarity in terms of simple form and distribution of colors. Lastly some of the confusions are completely acceptable based on the origins of these styles (art movements) that are noted in art history literature. For example,  
``Renaissance"(column 18) and ``Early Renaissance"(row 9); ``Post Impressionism" (column 21) and ``Impressionism"(row 13); ``Cubism" (8th row) and ``Synthetic Cubism" (column 26). Synthetic cubism is the later act of cubism with more color continued usage of collage and pasted papers, but less linear perspective than cubism.

\subsubsection{Genre Classification}
We narrowed down the list of all genres in our dataset (45 in total) to get a reasonable number of samples for each genre (10 selected genres are listed in table~\ref{fig:table_names}). We trained ten one-vs-all SVM  classifiers and compare their performance in Table~\ref{tab:genre}. In this table columns represent different features and rows are different metric that we used to compute the distance. As table~\ref{tab:genre} shows we achieved the best performance for genre classification by learning Boost metric on top of Classeme features. Generally the performance of these classifiers are better than classifiers that we trained for style classification. This is expected as the number of genres is less than the number of styles in our dataset.

Figure~\ref{fig:conf_genre_new} shows the confusion matrix for classification of genre by learning Boost metric, when we used Classeme features. Investigating the confusions that we find in this matrix, reveals interesting results. For example, our system confuses ``Landscape" (5th row) with ``Cityspace" (2nd column) and ``Genre paintings" (3rd column). However, this confusion is expected as art historians can find common elements in these genres. On one hand ``Landscape" paintings usually show rivers, mountains and valleys and there is no significant figure in them; frequently very similar to ``Genre paintings" as they capture daily life. The difference appears in the fact that despite the ``Genre paintings", ``Landscape" paintings are idealized.  On the other hand, ``Landscape" and ``Cityspace" paintings are  very similar as both have open space and use realistic color tonalities.

\begin{table}[t]
\begin{center}
\begin{tabular}{|l|l|l|l|l|l|}
\hline
 Metric~/~Features & GIST & Classemes & Picodes & CNN & Dim.\\
 \hline \hline
  Baseline & 17.58 & 45.29 & 45.82 & 20.38 & 512\\
  Boost & \textbf{25.65}  & \textbf{57.76} & \textbf{55.50} & 29.65 & 512\\
  ITML & 19.95 & 51.79 & 53.93 & \textbf{31.04} & 512\\	
  LMNN & 20.41 & 53.99 & 53.92 & 30.92 & 100\\
  MLKR & 21.22 & 49.61 & 19.54 & 21.77 & 512\\
  NCA & 18.80 & 53.70 & 53.81 & 22.26 & 23\\
\hline 
\end{tabular}
\caption{Accuracy for the task of artist classification}
\label{tab:artist}
\end{center}
\vspace*{-30pt}
\end{table}

\subsubsection{Artist Classification}
For the task of the artist classification, we trained one-vs-all SVM classifiers for each of 23 artists. For each test image, we determine its artist by finding the classifier that produces the maximum confidence. Table~\ref{tab:artist} shows the performance of different combinations of features and metrics for this task. In general learning Boost metric improves artist classification better than all other metrics, except the case of CNN features where learning ITML metric gained the best performance. We plotted the confusion matrix of this classification task in figure~\ref{fig:conf_artist_new}. In this plot, some confusions between artists are clearly reasonable. We investigated two cases: 

First case, ``Claude Monet"(5th row) and ``Camille Pissaro"(3rd column). Both of these Impressionist artists who lived in the late nineteen and early twentieth centuries. Interestingly, based on art history literature Monet and Pissaro became friends when they both attended the "Acad\'emie Suisse" in Paris. This friendship lasted for a long time and resulted in some noticeable interactions between them. Second case, paintings of ``Childe Hassam"(4th row) are mostly confused with ones from ``Monet"(5th column). This confusion is acceptable as Hassam is an American Impressionist, who declared himself as being influenced by French Impressionists. Hassam called himself an ``Extreme Impressionist", who painted some flag-themed artworks similar to Monet.

By looking at reported performances in tables~\ref{tab:style}-~\ref{tab:artist}, we conclude that, all three classification tasks can benefit from learning the appropriate metric. This means that we can improve the accuracy of baseline classification by learning metrics independent of the type of visual feature or the concept that we are classifying painting based on. Experimental results show that, independent of the task, NCA and MLKR approaches are performing worse than other metrics. Additionally, Boost metric always gives the best or the second best results for all classification tasks.

Regarding analysis of importance of features, we can verify that Classeme and Picode features are better image representations for classification purposes. Based on these classification experiments, we claim that Classemes and Picodes features perform better than CNN features. This is rooted in the fact that amount of supervision for training Classeme and Picodes is more than CNN training. Also, unlike Classeme and Picodes, CNN feature is designed to categorize the object insides a given bounding box. However, in the case of paintings we cannot assume that all the bounding boxes around the objects are given.

\begin{table}[t]
\begin{center}
\begin{tabular}{|l|l|l|l|l|}
\hline
 Task~/~Features & GIST & Classemes & Picodes & CNN \\
 \hline \hline
  Style & 20.21 & \textbf{37.33} & \textbf{33.27} & 21.99 \\
  Genre & 35.94 & \textbf{58.29} & \textbf{56.09} & 47.05\\
  Artist& 30.37	& \textbf{59.37} & \textbf{55.65} & 33.62\\
 \hline 
\end{tabular}
\caption{Classification performance for metric fusion methodology.}
\label{tab:metric_comb}
\end{center}
\vspace*{-30pt}
\end{table}

\subsubsection{Integration of Features and Metrics}
We investigated the performance of different metric learning approaches and visual features individually. In the next step, we find out the best performance for aforementioned classification tasks by combining different visual features. Toward this goal, we followed two strategies. First, for a given metric, we project visual features by applying the metric and concatenate these projected visual features together. Second, we fixed the type of visual feature that we use and project it with the application of different metrics and concatenate these projections all together. Having this larger feature vectors (either of two strategies), we train SVM classifiers for three tasks of Style, Genre and Artist classification. Table~\ref{tab:feat_comb} shows the results of these experiments where we followed the earlier strategy and table~\ref{tab:metric_comb} shows the results of the later case. In general we get better results by fixing the metric and concatenating the projected feature vectors (first strategy).

The work of Bar et al~\cite{israel} is the most similar to ours and we compare our final results of these experiments with their reported performance. \cite{israel} only performed the task of style classification on half of the images in our dataset and achieved the accuracy of 43\% by using two variations of PiCoDes features and two layers of CNN. However we outperform their approach by achieving 45.97 \% accuracy for the task of style classification when we used LMNN metric to project GIST, Classeme, PiCoDes and CNN features and concatenate them all together as it is reported in the third column of table~\ref{tab:feat_comb}.

Our contribution goes beyond outperforming state-of-the-art by learning a more compact feature representation. In this work, our best performance for style classification happens when we concatenate four 100-dimensional feature vectors. This results in a 400 dimensional feature vectors that we train SVM classifiers on top of them. However \cite{israel} extract a 3882 dimensional feature vector to their best reported performance. As a result we not only outperform the state-of-the-art, but presented a better image representation that reduces the amount of space by 90\%. Our efficient feature vector is an extremely useful image representation that gains the best classification accuracy and we consider its application for the task of image retrieval as future work. 

\begin{figure*}[t]
\includegraphics[width=\textwidth, height=.95\textheight]{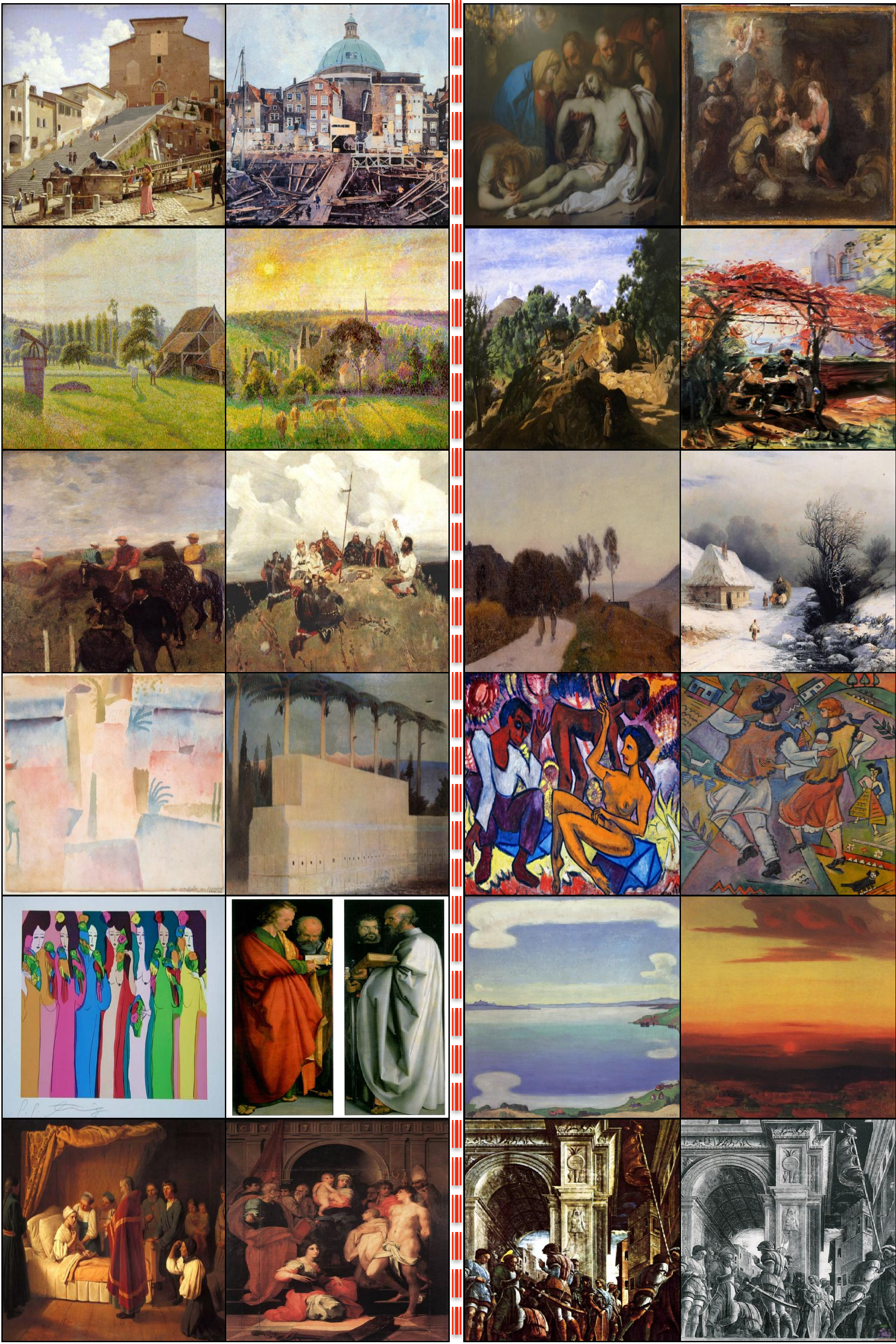}
\caption{Sample output for the tasks of image search. In each pair, the image on the left is the query and image on the right is the closest match, but not from the same style (LMNN plus feature fusion.)}
\label{fig:img_search}
\end{figure*}

To qualitatively evaluate extracted visual features and learned metrics, we did a prototype image search task. As the feature fusion with application of LMNN metric gives the best performance for style classification, we used this setting as our similarity measurement model. Figure~\ref{fig:img_search} shows some sample output of this image search task. For each pair, the image on the left is the query image, which we find the closest match(image on the right) to it based on LMNN and feature fusion. However we force the system to pick the closest match that does not belong to the same style as the query image. This verifies that although we learn the metric based on style labels, the learned projection can find similarity across styles.

\begin{table}[t]
\begin{center}
\begin{tabular}{|l|l|l|l|l|l|}
\hline
 Concept~/~Metric & Boost & ITML & LMNN & MKLR & NCA\\
 \hline \hline
  Style & 41.74 & \textbf{45.05}	& \textbf{45.97}	& 38.91	& 40.61\\
  Genre & \textbf{58.51}	& \textbf{60.28}	& 58.48	& 55.79 & 54.82\\
  Artist & \textbf{61.24}	 & 60.46 &	\textbf{63.06} & 53.19 &	55.83\\
\hline 
\end{tabular}
\caption{Classification results for feature fusion methodology.}
\label{tab:feat_comb}
\end{center}
\vspace*{-30pt}
\end{table}

\begin{table*}[t]
\includegraphics[width=\textwidth, height=.9\textheight]{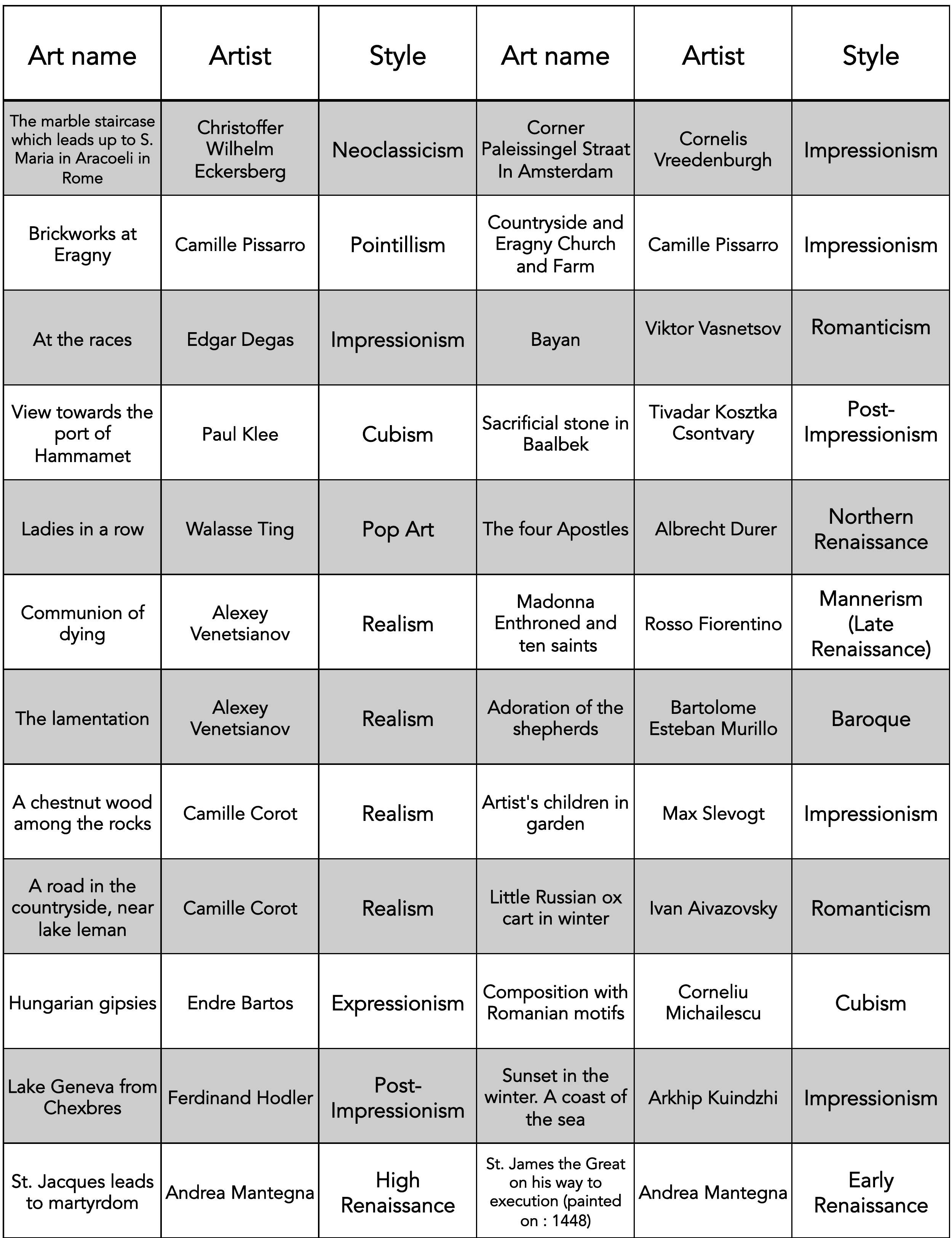}
\caption{Annotation of paintings in Figure~\ref{fig:img_search}. Each row corresponds to one pair of images, labeled with the name of painting, its style and its artist. First six rows correspond to the six pairs on the left in Figure~\ref{fig:img_search} and next six rows correspond to the pairs on the right.}
\label{tab:img_search_anno}
\end{table*}

\section{Conclusion and Future Works}
\label{Sec:con}
\label{Sec:con}
In this paper we investigated the applicability of metric learning approaches and performance of different visual features for learning similarity in a collection of fine-art paintings. We implemented meaningful metrics for measuring similarity between paintings. These metrics are learned in a supervised manner to put paintings from one concept close to each other and far from others. In this work we used three concepts: Style, Genre and Artist. We used these learned metrics to transform raw visual features into another space that we can significantly improve the performance for three important tasks of \textit{Style, Genre} and \textit{Artist classification}. We conducted our comparative experiments on the largest publicly available dataset of fine-art paintings to evaluate the performance for the aforementioned tasks. 

We conclude that:
\begin{itemize}
\item Classeme features show the superior performance for all three tasks of Style, Genre or Artist classification. This superior performance is independent of the type of metric that has been learned. 

\item In the case of working on individual type of visual features, Boost metric and Information Theoretic Metric Learning(ITML) approaches improve the accuracy of classification tasks across all features.

\item For the case of using different types of features all together(feature fusion), Large-Margin Nearest-Neighbor(LMNN) metric learning achieves the best performance for all classification experiments.

\item By learning LMNN metric on Classeme features, we find an optimized representation that not only outperforms state-of-the art for the task of style classification, but reduce the size of feature vector by 90\%.

\end{itemize} 

We consider verification of  applicability of this representation for the task of image retrieval and recommendation systems as future work. As other future works we would like to learn metrics based on other annotation(e.g. time period).


\bibliographystyle{abbrv}
\bibliography{MM2015-arXiv}

\begin{thebibliography}{10}

\bibitem{csift}
A.~E. Abdel-Hakim and A.~A. Farag.
\newblock Csift: A sift descriptor with color invariant characteristics.
\newblock In {\em IEEE Conference on Computer Vision and Pattern Recognition,
  CVPR}, 2006.

\bibitem{arnheim1969visual}
R.~Arnheim.
\newblock {\em Visual thinking}.
\newblock Univ of California Press, 1969.

\bibitem{Arora12}
R.~S. Arora and A.~M. Elgammal.
\newblock Towards automated classification of fine-art painting style: A
  comparative study.
\newblock In {\em ICPR}, 2012.

\bibitem{israel}
Y.~Bar, N.~Levy, and L.~Wolf.
\newblock Classification of artistic styles using binarized features derived
  from a deep neural network.
\newblock 2014.

\bibitem{stork_book}
A.~Bentkowska-Kafel and J.~Coddington.
\newblock {\em Computer Vision and Image Analysis of Art: Proceedings of the
  SPIE Electronic Imaging Symposium, San Jose Convention Center, 18-22 January
  2010}.
\newblock PROCEEDINGS OF SPIE. 2010.

\bibitem{berezhnoy2009automatic}
I.~E. Berezhnoy, E.~O. Postma, and H.~J. van~den Herik.
\newblock Automatic extraction of brushstroke orientation from paintings.
\newblock {\em Machine Vision and Applications}, 20(1):1--9, 2009.

\bibitem{classemes}
A.~Bergamo and L.~Torresani.
\newblock Classemes and other classifier-based features for efficient object
  categorization.
\newblock {\em IEEE Transactions on Pattern Analysis and Machine Intelligence},
  page~1, 2014.

\bibitem{picodes}
A.~Bergamo, L.~Torresani, and A.~W. Fitzgibbon.
\newblock Picodes: Learning a compact code for novel-category recognition.
\newblock In {\em Advances in Neural Information Processing Systems}, pages
  2088--2096, 2011.

\bibitem{Carneiro12}
G.~Carneiro, N.~P. da~Silva, A.~D. Bue, and J.~P. Costeira.
\newblock Artistic image classification: An analysis on the printart database.
\newblock In {\em ECCV}, 2012.

\bibitem{HOG}
N.~Dalal and B.~Triggs.
\newblock Histograms of oriented gradients for human detection.
\newblock In {\em International Conference on Computer Vision \& Pattern
  Recognition}, volume~2, pages 886--893, June 2005.

\bibitem{ITML}
J.~V. Davis, B.~Kulis, P.~Jain, S.~Sra, and I.~S. Dhillon.
\newblock Information-theoretic metric learning.
\newblock In {\em ICML}, 2007.

\bibitem{khan}
M.~V. Fahad Shahbaz~Khan, Joost van de~Weijer.
\newblock Who painted this painting?
\newblock 2010.

\bibitem{lois}
L.~Fichner-Rathus.
\newblock {\em Foundations of Art and Design}.
\newblock Clark Baxter, 2008.

\bibitem{NCA}
J.~Goldberger, S.~Roweis, G.~Hinton, and R.~Salakhutdinov.
\newblock Neighbourhood components analysis.
\newblock In {\em NIPS}, 2004.

\bibitem{johnson2008image}
C.~R. Johnson, E.~Hendriks, I.~J. Berezhnoy, E.~Brevdo, S.~M. Hughes,
  I.~Daubechies, J.~Li, E.~Postma, and J.~Z. Wang.
\newblock Image processing for artist identification.
\newblock {\em Signal Processing Magazine, IEEE}, 25(4):37--48, 2008.

\bibitem{alex}
A.~Krizhevsky, I.~Sutskever, and G.~E. Hinton.
\newblock Imagenet classification with deep convolutional neural networks.
\newblock In {\em Advances in neural information processing systems}, pages
  1097--1105, 2012.

\bibitem{lecun1998gradient}
Y.~LeCun, L.~Bottou, Y.~Bengio, and P.~Haffner.
\newblock Gradient-based learning applied to document recognition.
\newblock {\em Proceedings of the IEEE}, 86(11):2278--2324, 1998.

\bibitem{li2004studying}
J.~Li and J.~Z. Wang.
\newblock Studying digital imagery of ancient paintings by mixtures of
  stochastic models.
\newblock {\em Image Processing, IEEE Transactions on}, 13(3):340--353, 2004.

\bibitem{Jia12}
J.~Li, L.~Yao, E.~Hendriks, and J.~Z. Wang.
\newblock Rhythmic brushstrokes distinguish van gogh from his contemporaries:
  Findings via automated brushstroke extraction.
\newblock {\em IEEE Trans. Pattern Anal. Mach. Intell.}, 2012.

\bibitem{Lombardi}
T.~E. Lombardi.
\newblock The classification of style in fine-art painting.
\newblock {\em ETD Collection for Pace University. Paper AAI3189084.}, 2005.

\bibitem{SIFT}
D.~G. Lowe.
\newblock Distinctive image features from scale-invariant keypoints.
\newblock {\em Int. J. Comput. Vision}, 2004.

\bibitem{lyu2004digital}
S.~Lyu, D.~Rockmore, and H.~Farid.
\newblock A digital technique for art authentication.
\newblock {\em Proceedings of the National Academy of Sciences of the United
  States of America}, 101(49):17006--17010, 2004.

\bibitem{GIST}
A.~Oliva and A.~Torralba.
\newblock Modeling the shape of the scene: A holistic representation of the
  spatial envelope.
\newblock {\em IJCV}, 2001.

\bibitem{brdahujapo09}
G.~Polatkan, S.~Jafarpour, A.~Brasoveanu, S.~Hughes, and I.~Daubechies.
\newblock Detection of forgery in paintings using supervised learning.
\newblock In {\em 16th IEEE International Conference on Image Processing
  (ICIP)}, 2009.

\bibitem{sablatnig}
R.~Sablatnig, P.~Kammerer, and E.~Zolda.
\newblock Hierarchical classification of paintings using face- and brush stroke
  models.
\newblock 1998.

\bibitem{saleh2014knowledge}
B.~Saleh, K.~Abe, and A.~Elgammal.
\newblock Knowledge discovery of artistic influences: A metric learning
  approach.
\newblock In {\em ICCC}, 2014.

\bibitem{boostmetric}
C.~Shen, J.~Kim, L.~Wang, and A.~van~den Hengel.
\newblock Positive semidefinite metric learning using boosting-like algorithms.
\newblock {\em Journal of Machine Learning Research}, 13:1007--1036, 2012.

\bibitem{stork2009computer}
D.~G. Stork.
\newblock Computer vision and computer graphics analysis of paintings and
  drawings: An introduction to the literature.
\newblock In {\em Computer Analysis of Images and Patterns}, pages 9--24.
  Springer, 2009.

\bibitem{aleb}
L.~Torresani, M.~Szummer, and A.~Fitzgibbon.
\newblock Efficient object category recognition using classemes.
\newblock In {\em ECCV}, 2010.

\bibitem{matconv}
A.~Vedaldi and K.~Lenc.
\newblock Matconvnet -- convolutional neural networks for matlab.
\newblock {\em CoRR}, abs/1412.4564, 2014.

\bibitem{MLKR}
K.~Weinberger and G.~Tesauro.
\newblock Metric learning for kernel regression.
\newblock In {\em Eleventh international conference on artificial intelligence
  and statistics}, pages 608--615, 2007.

\bibitem{LMNN}
K.~Q. Weinberger and L.~K. Saul.
\newblock Distance metric learning for large margin nearest neighbor
  classification.
\newblock {\em JMLR}, 2009.

\end{thebibliography}

\end{document}